%% file: icon2014.tex
\documentclass[11pt]{article}
\usepackage{acl2016}
\usepackage{times}
\usepackage[hyphens]{url}
\usepackage{latexsym}
\usepackage{xspace}
\usepackage[utf8]{inputenc}
\usepackage{amsmath}

\usepackage{algorithm}
\usepackage{algpseudocode}

\makeatletter
\def\BState{\State\hskip-\ALG@thistlm}
\makeatother

\pdfoutput=1
\aclfinalcopy 

\title{Statistical Machine Translation for Indian Languages: Mission Hindi}

\author{Raj Nath Patel \\
  KBCS, CDAC Mumbai \\
  {\tt rajnathp@cdac.in} \\\And
  Prakash B. Pimpale \\
  KBCS, CDAC Mumbai \\
  {\tt prakash@cdac.in} \\\And
  Sasikumar M. \\
  KBCS, CDAC Mumbai \\
  {\tt sasi@cdac.in} \\}

\date{}

\newcommand{\bnhi}{\textit{be-hi}\xspace}
\newcommand{\enhi}{\textit{en-hi}\xspace}
\newcommand{\tahi}{\textit{ta-hi}\xspace}
\newcommand{\tehi}{\textit{te-hi}\xspace}
\newcommand{\mrhi}{\textit{mr-hi}\xspace}

\begin{document}
\maketitle
\begin{abstract}
This paper discusses Centre for Development of Advanced Computing Mumbai's (CDACM) submission to the NLP Tools Contest on Statistical Machine Translation in Indian Languages (ILSMT) 2014 (collocated with ICON 2014). The objective of the contest was to explore the effectiveness of Statistical Machine Translation (SMT) for Indian language to Indian language
and English-Hindi machine translation. In this paper, we have proposed that suffix separation and word splitting for SMT from agglutinative languages to Hindi significantly improves over the baseline (BL). We have also shown that the factored model with reordering outperforms the phrase-based SMT for English-Hindi (\enhi). We report our work on all five pairs of languages, namely Bengali-Hindi (\bnhi), Marathi-Hindi (\mrhi), Tamil-Hindi (\tahi), Telugu-Hindi (\tehi), and \enhi for Health, Tourism, and General domains.
\end{abstract}

\input{intro}
\input{methodoogy}

\input{settings}
\input{results}
\input{submission}

\input{analysis}
\input{conclusion}

\bibliography{acl2016}
\bibliographystyle{acl2016}
\end{document}

%% file: intro.tex

\section{Introduction}
\label{sec:intro}
In this paper, we present our experiments on SMT from Bengali, Marathi, Tamil, Telugu and English to Hindi. From the set of languages involved in the shared task, Bengali, Hindi and Marathi belong to IndoAryan family and Tamil and Telugu are from Dravidian language family. All languages except English, have the same flexibility towards word order, canonically following the SOV structure.

With reference to the morphology, Bengali, Marathi, Tamil, and Telugu are more agglutinative compared to Hindi. It is known that SMT produces more unknown words resulting in the bad translation quality if the morphological divergence between the source and target language is high. ~\newcite{koehn:2003:CS},~\newcite{popovic:2004:word-stem} and~\newcite{popovic:2006:CW} have demonstrated ways to handle this issue with morphological segmentation of words before training the SMT system. To tackle the morphological divergence of Hindi with these languages we have explored Suffix Separation (SS) and Compound word Splitting (CS) as a pre-processing step.

For English to Hindi SMT, better alignment is achieved through the use of preordering developed by~\newcite{patel:2013} and  stem as an alignment factor~\cite{koehn:2007:factored}. The rest of the paper is organized as follows. In Section~\ref{sec:method}, we discuss our methodology, followed by data-set and experimental setup in section~\ref{sec:settings}. Section~\ref{sec:results} discusses experiments and results. Submitted systems to the shared task and error analysis are displayed in section~\ref{sec:submit} and~\ref{sec:analysis} respectively, followed by conclusion and future work in section~\ref{sec:conclusion}.

%% file: methodoogy.tex

\section{Methodology}
\label{sec:method}
Our methodology to tackle morphological and structural divergence involves the use of the suffix separation, compound splitting, and reordering for the language pairs under study. These methods are briefly described below. Pseudocode for the suffix separation and compound word splitting is detailed in Algorithms~\ref{algo:ss} and~\ref{algo:cs} respectively.

\begin{algorithm*} \small
	\caption{Suffix Separation}\label{algo:ss}
	\begin{algorithmic}[1]
		\Procedure{SuffixSep}{$word$}
		\State $\textit{suffixSet} \gets \text{read file } \textit{suffix list}$
		\State $\textit{splits} \gets \textit{\{word, “NULL”\}}$
		\For{$\text{suffix} \gets \text{suffixSet}$}
			\If $\textit{ word.ENDSWITH} = \text{suffix} \And \textit{word.LENGTH} > \textit{suffix.LENGTH}$
			\State $\text{splits[0]} \gets \textit{word.SUBSTRING(0, word.LASTINDEXOF(suffix))}$
			\State $\text{splits[1]} \gets \textit{suffix}$
			\Return $splits$
			\EndIf
		\EndFor
		\EndProcedure
	\end{algorithmic}
\end{algorithm*}

\begin{algorithm*} \small
	\caption{Compound Splitting}\label{algo:cs}
	\begin{algorithmic}[1]
		\Procedure{CompoundSplit}{}
		\State $\textit{vocab} \gets \text{read monoligual corpus } \textit{unique word list}$
		\For{$\text{word} \gets \textit{file}$}
			\For{$\text{vcb} \gets \textit{vocab}$}
				\If $\text{ word.ENDSWITH} =  \text{vcb} \And \text{word.LENGTH} > \text{vcb.LENGTH+5}$
				\State $\text{compoundSuffix} \gets \textit{vcb}$
				\State $\text{Delete vcb for vocab}$
				\Return $\textit{compoundSuffix}$
				\EndIf
			\EndFor
		\EndFor
		\EndProcedure
	\end{algorithmic}
\end{algorithm*}

\subsection{Suffix Separation (SS)}
\label{subsec:ss}
In this step, the source language words are pre-processed for suffix separation. We have considered only suffix from source language which corresponds to post-positions in Hindi. For example, in Marathi, \footnotemark '\textbf{\textit{mahinyaaMnii}}' \footnotetext{All \textbf{\textit{Non-English} (Marathi and Hindi)} words have been written in Itrans using~\url{http://sanskritlibrary.org/transcodeText.html}} is translated as '\textbf{\textit{mahiine meM}}' in Hindi. In this case, we split the word into '\textbf{\textit{mahiny + aaMnii}}'. Here, the suffix '\textbf{\textit{aaMnii}}' corresponds to the word '\textbf{\textit{meM}}' in Hindi. For this task, the list of suffixes is manually created with the linguistic expertise. When a word is subjected to SS, longest matching suffix from the list is considered for suffix separation. Suffix separation takes place only once for a word.

\subsection{Compound Splitting (CS)}
\label{subsec:cs}
In this step source language compound words are split into constituents, recursively. For example, in Marathi, a compound word '\textbf{\textit{daMtatajGYaaMkaDuuna}}' is translated as '\textbf{\textit{danta visheShaGYa se}}' in Hindi. In this case we split the source word into constituents, '\textbf{\textit{daMta}}' , '\textbf{\textit{tajGYaaM}}' and '\textbf{\textit{kaDuuna}}'. The list of the constituent suffixes for splitting is empirically prepared from a monolingual data. The compound suffix creation algorithm is very basic and simple, the pseudo code is detailed in Algoritm~\ref{algo:cs}. A word is considered for compound suffix list, if it appears as a suffix in another word of the monolingual corpus.

\subsection{Reordering (RO)}
\label{subsec:ro}
It is based on the syntactic transformation of the English sentence parse tree according to the target language (Hindi) structure. We have used source side reordering developed by~\newcite{patel:2013}, and~\newcite{ramanathan:2008}.

%% file: settings.tex

\begin{table}[!hbt] \small
	\centering
	\begin{tabular}{l|c|c|c}
		& health & tourism & general \\ \hline
		training (TM) & 24K & 24K & 48K \\
		training (LM) & 71K & 71K & 71K \\
		test1 & 500 & 500 & 1000 \\
		test2 & 500 & 500 & 1000 \\
	\end{tabular}
	\caption{Sentence count for training, testing and development data; TM: Translation Model, LM: Language Model}
	\label{tab:data}
\end{table}

\section{Data-set and Experimental Setup}
\label{sec:settings}
We now discuss training and testing corpus from health, tourism and general domains for \bnhi, \mrhi, \tahi, \tehi, and \enhi language pairs, followed by preprocessing, SMT system setup and evaluation
metrics for experiments.

\subsection{Corpus for SMT Training and Testing}
\label{subsec:data}
For experiments, we have used corpus shared by ILSMT organizers, detailed in Table~\ref{tab:data}. Additional monolingual corpus of approx., 23K sentences~\cite{khapra:2010} is used to train the language model. The evaluation of the systems were done using Test1 data which was the development set. The quality of the submitted systems were estimated by the organizers against Test2 corpus.

\subsection{Pre-Processing}
\label{subsec:per}
To tackle the morphological divergence between the source and target languages (Bengali/Marathi/Tamil/Telugu to Hindi), we used suffix separation and compound splitting, as explained in section~\ref{sec:method}. To handle the structural divergence for English-Hindi SMT, we exploited source side preordering~\cite{patel:2013,ramanathan:2008}.

\subsection{SMT System Set up}
\label{subsec:setup}
The baseline system was setup using the phrase-based model (~\cite{brown:1990,marcu:2002,och:2003,koehn:2003} and~\newcite{koehn:2007} was used for factored model. The language model was trained using KenLM~\cite{heafield:2011} toolkit with modified Kneser-Ney smoothing~\cite{chen:1996}. For factored SMT training, we used source and target side stem as an alignment factor. Stemming was done using lightweight stemmer~\cite{ramanathan:2003:stem} for Hindi. For English, we used porter stemmer~\cite{minnen:2001}.

\subsection{Evaluation Metrics}
\label{subsec:eval}
We compared different experimental systems using BLEU~\cite{papineni:2002}, NIST~\cite{doddington:2002} and translation edit rate (TER)~\cite{snover:2006:ter}. For a MT system to be better, higher BLEU and NIST scores with lower TER are desired.

%% file: results.tex

\section{Experiments and Results}
\label{sec:results}
In the following subsections, we present different experiments carried out for the shared task. We also study the impact of suffix separation, compound splitting and preordering on the SMT accuracy.

\subsection{Impact of Suffix Separation and Compound Splitting}
\label{subsec:results:sscs}


\begin{table*}[!hbt] \small
	\centering
	\fontsize{7pt}{0}
	\begin{tabular}{|p{1.5cm}|c|c|c|p{2.5cm}|c|c|c|c|c|}
		\hline
		SRC sentence (mr) & \textit{dara} & \textit{sahaa} & \textit{mahinyaaMnii} & \textit{daMtatajGYaaMkaDuuna} & \textit{tapaasuuna} & \textit{ghyaa} & & & \\
		BL aligned (hi) & \textit{hara} & \textit{Chaha} & \textit{mahiine meM} & \textit{danta visheShaGYa se chekaapa karaaeM} & - & - & & & \\ \hline
		SRC' Sentence (mr) & \textit{dara} & \textit{sahaa} & \textit{mahiny} & \textit{aaMnii} & \textit{daMta} & \textit{tajGYaaM} & \textit{kaDuuna} & \textit{tapaasuuna} &  \textit{ghyaa} \\
		BL+CS+SS aligned (hi) & \textit{hara} & \textit{Chaha} & \textit{mahiine} & \textit{meM} & \textit{danta} & \textit{visheShaGYa} & \textit{se} & \textit{chekaapa} & \textit{karaaeM} \\ \hline
	\end{tabular}
	\caption{Improvement in alignments; SRC: Source, SRC': Pre-processed Source}
	\label{tab:alignments}
\end{table*}

Pre-processing of source words for suffix separation and compound word splitting results in better alignment and hence the better translation. The alignment improvements can be seen in the Table~\ref{tab:alignments}. Improvement in the translation quality can be observed in the various evaluation scores detailed in Table~\ref{tab:CSSS}. From the table, we can infer that for \bnhi and \mrhi, suffix separation have shown significant improvements over the baseline, whereas, compound word splitting has caused slight improvement. However, compound splitting has found to be more effective than suffix separation for \tahi and \tehi.

We have also tried a combination of compound word splitting and suffix separation, serially. We observed improvements over the BL+SS and BL+CS for \tahi and \tehi, across the evaluation scores. BL+SS is better than all other systems for \bnhi, across the evaluation metrics, whereas, for \mrhi BL+CS+SS has highest BLEU and BL+SS has the best NIST and TER scores. Compound splitting for Bengali and Marathi can be further investigated to improve the systems.

\begin{table}[!hbt] \small
	\centering
	\begin{tabular}{l|l|c|c|c}
		& models & BLEU & NIST & TER \\ \hline
	\bnhi & BL & 30.16 & 6.906 & 48.26 \\
		& BL+SS & \bf 31.73 & \bf 6.993 & \bf 46.92  \\
		& BL+CS & 30.33 & 6.792 &  49.08 \\
		& BL+CS+SS & 30.34 & 6.804 & 48.58 \\ \hline
	\mrhi & BL & 35.56 & 7.478 & 42.98 \\
		& BL+SS & 39.84 & \bf 7.877 & \bf 39.44 \\
		& BL+CS & 37.87 & 7.502 & 43.01 \\
		& BL+CS+SS & \bf 39.88 & 7.796 & 39.93 \\ \hline
	\tehi & BL & 16.76 & 4.772 & 64.48 \\
		& BL+SS & 17.49 & 4.927 & 63.76 \\
		& BL+CS & 19.82 & 5.260 & 62.96 \\
		& BL+CS+SS & \bf 20.16 & \bf 5.300 & \bf 62.69 \\ \hline
	\tahi & BL & 24.89 & 6.205 & 52.31 \\
		& BL+SS & 27.18 & 6.473 & 50.39 \\
		& BL+CS & 27.47 & 6.448 & 51.01 \\
		& BL+CS+SS & \bf 28.95 & \bf 6.535 & \bf 49.95
	\end{tabular}
	\caption{Effect of suffix separation and compound splitting}
	\label{tab:CSSS}
\end{table}

\subsection{Impact of Reordering}
\label{subsec:results:ro}
Preordering of the source language sentence helps in the better alignment and decoding for English to Indian language~\cite{ramanathan:2008,patel:2013,kunchukuttan:2014} SMT. Table~\ref{tab:RO} details the results for the systems under study. We can see that BL+RO shows significant improvement over BL. Further, the factored SMT system with stem as alignment factor shows slight improvement in BLEU over the BL+RO, but other metrics show BL+RO is better
compared to the factored system.

\begin{table}[!hbt] \small
	\centering
	\begin{tabular}{l|l|c|c|c}
		& models & BLEU & NIST & TER \\ \hline
		\enhi & BL & 18.52 & 5.813 & 64.33 \\
		& BL+RO & 22.72 & \bf 6.035 & \bf 59.85 \\
		& BL+RO+FACT & \bf 22.83 & 5.994 & 60.17
	\end{tabular}
	\caption{Effect of source reordering (RO) and factor (FACT)}
	\label{tab:RO}
\end{table}

%% file: submission.tex

\section{Submission}
\label{sec:submit}

\begin{table*}[!hbt] \small
	\centering
	\begin{tabular}{l|l|l|c|c|c|c|c|c|c|c|c}
		& & & \multicolumn{3}{c|}{health} & \multicolumn{3}{c|}{tourism} & \multicolumn{3}{c}{general} \\ \hline
		& & & \multicolumn{1}{c|}{BLEU} & \multicolumn{1}{c|}{NIST} & \multicolumn{1}{c|}{TER} & \multicolumn{1}{c|}{BLEU} & \multicolumn{1}{c|}{NIST} & \multicolumn{1}{c|}{TER} & \multicolumn{1}{c|}{BLEU} & \multicolumn{1}{c|}{NIST} & \multicolumn{1}{c}{TER} \\ \hline
	\bnhi & BL & T1 & 30.78 & 6.628 & 47.68 & 29.35 & 6.389 & 48.95 & 30.16 & 6.906 & 48.26 \\
		& FSS & T1 & 31.95 & 6.746 & 45.98 & 30.76 & 6.433 & 47.98 & 31.73 & 6.993 & 46.92 \\ 
		& & T2 & \bf 31.40 & 6.723 & \bf 45.55 & \bf 31.17 & 6.605 & \bf 46.25 & \bf 31.34 & 7.079 & \bf 45.98 \\ \hline
	\mrhi & BL & T1 & 35.42 & 7.103 & 42.52 & 35.20 & 6.917 & 43.81 & 35.56 & 7.478 & 42.98 \\
		& FSS & T1 & 39.03 & 7.390 & 39.85 & 39.45 & 7.148 & 41.06 & 39.88 & 7.796 & 39.93 \\ 
		& & T2 & \bf 39.31 & 7.192 & \bf 41.85 & \bf 37.73 & 7.199 & \bf 41.25 & \bf 39.10 & 7.718 & \bf 41.02 \\ \hline
	\tahi & BL & T1 & 17.99 & 4.646 & 63.79 & 14.44 & 4.216 & 65.90 & 16.76 & 4.772 & 64.48 \\
		& FSS & T1 & 20.83 & 5.104 & 62.18 & 17.86 & 4.682 & 64.66 & 20.16 & 5.300 & 62.69 \\ 
		& & T2 & \bf 20.69 & 5.075 & \bf 62.59 & \bf 17.07 & 4.714 & \bf 63.78 & \bf 19.81 & 5.247 & \bf 62.33 \\ \hline
	\tehi & BL & T1 & 24.83 & 5.856 & 52.25 & 23.85 & 5.679 & 53.57 & 24.89 & 6.205 & 52.31 \\
		& FSS & T1 & 29.90 & 6.322 & 49.41 & 27.45 & 5.934 & 51.69 & 28.95 & 6.535 & 49.95 \\ 
		& & T2 & \bf 30.38 & 6.373 & \bf 49.28 & \bf 26.21 & 5.862 & \bf 52.90 & \bf 28.95 & 6.569 & \bf 50.47 \\ \hline
 	\enhi & BL & T1 & 20.12 & 5.840 & 61.49 & 16.97 & 5.233 & 66.83 & 18.52 & 5.813 & 64.33 \\
		& FSS & T1 & 23.16 & 5.870 & 58.54 & 20.98 & 5.342 & 62.90 & 22.83 & 5.994 & 60.17 \\ 
		& & T2 & \bf 21.84 & 5.662 & \bf 60.60 & \bf 19.45 & 5.349 & \bf 62.95 & \bf 20.97 & 5.837 & \bf 61.56
	\end{tabular}
	\caption{Consolidated Results; FSS: Final Submitted System; T1: Test1; T2: Test2; \textbf{BOLD} scores were provided by the organizers}
	\label{tab:submit}
\end{table*}

In this section, we discuss evaluation scores of the submitted systems. We submitted BL+SS for \bnhi and BL+CS+SS for all other language pairs except \enhi. For \enhi BL+RO+FACT is submitted as the final system. Table~\ref{tab:submit} summarizes the evaluation of the submission against Test1 and Test2.

%% file: analysis.tex

\section{Error Analysis}
\label{sec:analysis}
A closer look at the performance of these systems to understand the utility of SS and CS has been done. We report a few early observations.

\subsection{Superfluous Splitting}
\label{subsec:analysis:ss}
With the suffix separation, the Marathi word like '\textbf{\textit{dilaavara}}' is getting split into '\textbf{\textit{dilaa}}' + '\textbf{\textit{vara}}' which is a wrong split. As, '\textbf{\textit{dilaavara}}' is a proper noun and hence should not have been split. We tried to overcome this error by avoiding suffix separation and compound splitting of NNP POS tagged words, but that was stopping many other valid candidates from pre-processing.

\subsection{Bad Split}
\label{subsec:analysis:bs}
The words like '\textbf{\textit{jarmaniitiila}}' is getting split into '\textbf{\textit{jarmaniit}} + '\textbf{\textit{iila}} which actually should have been split into '\textbf{\textit{jarmanii}} +'\textbf{\textit{tiila}}'. Similarly, many words on splitting have not given any valid Marathi word which causesed sparsity in training data up to some extent.

%% file: conclusion.tex

\section{Conclusion and Future Work}
\label{sec:conclusion}
In this paper, we presented various systems for translation from Bengali, English, Marathi, Tamil and Telugu to Hindi. These SMT systems with the use of source side suffix separation, compound splitting and preordering showed significantly higher accuracy over the baseline. In future, we could investigate the formulation of more effective solutions for SS, CS and RO. Reasons for lower BLEU due to the combination of suffix separation and compound word splitting for Bengali and Marathi is another interesting case to study further.